# Cross-lingual Transfer for Text Classification with Dictionary-based Heterogeneous Graph

**Nuttapong Chairatanakul[1,2], Noppayut Sriwatanasakdi[3], Nontawat Charoenphakdee[4,5]**
**Xin Liu[2,6], Tsuyoshi Murata[1,2]**

[1]Tokyo Institute of Technology, [2]RWBC-OIL, AIST, [3]Asurion Japan Holdings G.K.
[4]The University of Tokyo, [5]RIKEN AIP, [6]AIRC, AIST
`nuttapong.c@net.c.titech.ac.jp, noppayut.sriwatanasakdi@asurion.com`
`nontawat@ms.k.u-tokyo.ac.jp, xin.liu@aist.go.jp, murata@c.titech.ac.jp`

## Abstract

In cross-lingual text classification, it is required that *task-specific* training data in high-resource source languages are available, where the task is identical to that of a low-resource target language. However, collecting such training data can be infeasible because of the labeling cost, task characteristics, and privacy concerns. This paper proposes an alternative solution that uses only *task-independent* word embeddings of high-resource languages and bilingual dictionaries. First, we construct a dictionary-based heterogeneous graph (DHG) from bilingual dictionaries. This opens the possibility to use graph neural networks for cross-lingual transfer. The remaining challenge is the heterogeneity of DHG because multiple languages are considered. To address this challenge, we propose *dictionary-based heterogeneous graph neural network* (DHGNet) that effectively handles the heterogeneity of DHG by two-step aggregations, which are word-level and language-level aggregations. Experimental results demonstrate that our method outperforms pretrained models even though it does not access to large corpora. Furthermore, it can perform well even though dictionaries contain many incorrect translations. Its robustness allows the usage of a wider range of dictionaries such as an automatically constructed dictionary and crowdsourced dictionary, which are convenient for real-world applications.

## 1 Introduction

Modern machine learning methods typically require a large amount of data to achieve desirable performance (LeCun et al., 2015; Schmidhuber, 2015; Deng and Liu, 2018). While such a requirement can be feasible for languages such as English (Singh, 2008) (i.e., high-resource language), it is not the case for low-resource languages that lack sufficiently large corpora to build reliable statistical models (Cieri et al., 2016; Haffari et al., 2018). Because there are more than six thousand languages in the world (Nettle, 1998), and only a few of them are high-resource, it is important to enable the use of machine learning in low-resource languages. Cross-lingual text classification (CLTC) is a transfer learning paradigm that aims to incorporate the training data in high-resource languages (i.e., source languages) to solve the classification task in a low-resource language (i.e., target language) more effectively (Karamanolakis et al., 2020; Xu et al., 2016; Bel et al., 2003; Ruder, 2019).

In CLTC, it is common to assume that the source and target tasks are identical, and training labeled data in high-resource languages (i.e., source data) are available. Such assumptions can be restrictive, and we give the following four examples. First, source data may not be allowed to use owing to the data privacy, e.g., when customers disagree to disclose their opinion to public (Chidlovskii et al., 2016; Liang et al., 2020; Kundu et al., 2020). Second, we may not be able to keep the source data, which is also a motivation of sequential transfer learning (Ruder, 2019). The colossal-size data that are used to train BERT (Devlin et al., 2019) is a good example. The data of this caliber hardly fit in household PC storage. Third, it is possible that the target task is quite specific to the target language, which makes it difficult to find the source data in high-resource languages. One example is fake news classification. The news content is highly specific to each region, which may not be reported in a high-resource language. Fourth, collecting source labeled data usually requires the prior knowledge of that source language. For example, it is difficult for people who cannot speak Chinese to reliably collect data in Chinese language. For these reasons, it is beneficial to consider cross-lingual transfer where we do not require task-specific source data in high-resource languages.

Our goal is to overcome the unavailability of task-specific source data by enabling any low-resource languages to utilize high-quality and widely-available resources from any high-resource languages. To achieve this, we design a method that requires only task-independent word embeddings of a source language (e.g. French) and a word-level bilingual dictionary (e.g. French-Malayalam dictionary) to solve a classification task in the target language. Word embeddings can be easily obtained from many sources (Pennington et al., 2014; Mikolov et al., 2013a; Bojanowski et al., 2017). Likewise, bilingual dictionaries are available as man-made commercial products, free lexical database (Kamholz et al., 2014), or results from dictionary induction algorithms (Choe et al., 2020; Lample et al., 2018).

The main challenge of our problem is how to utilize source word embeddings and bilingual dictionaries, which is not straightforward for the following reasons. First, given a word, there are many choices of translation. Even worse, dictionaries may contain wrong translations, which is often the case for automatically constructed dictionaries. Moreover, the compatibility of the source and target languages can be different depending on the context. Finally, the quality of source embeddings and bilingual dictionaries can be diverse. This paper aims to design a machine learning method that effectively transfers task-independent embeddings of source languages to task-specific embeddings of the target language. The method should be able to determine the appropriate transfer for each word in the source languages to the translated words in the target language under such circumstances.

To solve this problem, we convert bilingual dictionaries into a *dictionary-based heterogeneous graph* (DHG) that represents words and translations as nodes and edges, respectively. This reduces the problem into graph representation learning, which allows us to use powerful methods, e.g., graph neural networks (GNNs), to solve. DHG is heterogeneous because there are many node types (languages) and edge types (language pairs), which are often ignored by GNNs in general (Wang et al., 2019; Hu et al., 2020; Chairatanakul et al., 2021). Then, to effectively address the heterogeneity of DHG, we propose dictionary-based heterogeneous graph neural networks (DHGNets) that first aggregate word translations for each language pair (word-level aggregation), and then aggregate the results from all languages (language-level aggregation).

Our contributions can be summarized as follows: First, we propose an alternative solution that enables cross-lingual transfer for text classification by using only 1) task-independent word embeddings of high-resource languages and 2) bilingual dictionaries. Second, we propose DHG, which can be utilized for cross-lingual transfer by any learning algorithms operating on graphs, such as GNNs. Third, we propose DHGNets, which effectively uses DHG to solve text classification in a low-resource language. Fourth, we provide experimental results to analyze and show the usefulness of the proposed solution and provide extensive analysis of the choice of dictionary, high-resource language, word embeddings, and GNNs. The code and resources are available at https://github.com/nutcrtnk/DHGNet.

## 2 Related Work

**Transfer learning** – The goal of transfer learning is to solve a target task with limited target data by incorporating source knowledge from other domains (Pan and Yang, 2009; Ruder, 2019). The challenge of this problem is how to make use of source knowledge and avoid *negative transfer*, which is a phenomenon where using source knowledge worsens the performance (Rosenstein et al., 2005). Our problem can be categorized as transfer learning where source knowledge does not include source labeled data but only bilingual dictionaries and task-independent word embeddings of high-resource languages. Recently, the transfer learning problem where no source labeled data are provided is called source-free domain adaptation and has been studied extensively in computer vision because of its practicality (Vongkulbhisal et al., 2019; Liang et al., 2020; Kundu et al., 2020). However, the study of this problem for cross-lingual transfer is limited to the best of our knowledge.

Note that our problem is significantly different from *zero-shot cross-lingual transfer*. In that problem, although target data are not available, it is often assumed that source labeled data and the additional target task information are available (Farhadi et al., 2009; Romera-Paredes and Torr, 2015; Phang et al., 2020). Furthermore, most work in NLP assumes that source and target domains either share the same task (Upadhyay et al., 2018; Liu et al., 2019, 2020) or language (Veeranna et al., 2016; Zhang et al., 2019).

**Cross-lingual text classification** – According to the transfer learning taxonomy proposed by Ruder (2019), CLTC is the most related problem to ours. However, CLTC assumes that source labeled data are available, and the source and target tasks are identical (Upadhyay et al., 2016; Conneau et al., 2018; Ruder et al., 2019; Karamanolakis et al., 2020; Xu et al., 2016; Bel et al., 2003). Since the data requirement of CLTC can be restrictive, there exist recent methods for weakly supervised CLTC where target labels are not required (Karamanolakis et al., 2020; Xu et al., 2016; Zhang et al., 2020a). Note that source labeled data are still required for such methods. To enable the use of cross-lingual transfer for more applications, our work explores a different direction where source labeled data are unavailable.

**Cross-lingual word embedding (CLWE)** – CLWEs are the representations of words that are typically learned by identifying mappings to map the monolingual word embeddings of each language to a shared embedding representations by utilizing resources such as aligned corpus and dictionary (Zhang et al., 2020b). Such representations are highly useful for comparing the meaning of words across languages (Xu et al., 2018; Vulić et al., 2019; Grave et al., 2019; Ruder et al., 2019). Learning CLWEs can also improve the performance of classification in a low-resource language (Duong et al., 2016; Vulić et al., 2019; Ruder et al., 2019; Zhang et al., 2020b).

**Pretrained model** – Pretraining methods have demonstrated their effectiveness in transfer learning for many NLP tasks. Pretraining typically requires a large amount of data, which can be unlabeled data, to effectively learn a good pretrained model for a general NLP task. Examples of methods in this family are BERT (Devlin et al., 2019), CoVe (McCann et al., 2017), ULMFiT (Howard and Ruder, 2018), and USE (Yang et al., 2020).

**Bilingual dictionary** – A bilingual dictionary maps words from a source to their translations in a target language. Its usage is found across cross-lingual tasks. Obviously in machine translation (Nießen and Ney, 2004; Duan et al., 2020; Zoph et al., 2016), dictionaries are used to provide ground-truth for word-level translation. Enhancing the quality of word embedding of low-resource languages is also possible with dictionary as a bridge to connect two languages (Duong et al., 2016; Mikolov et al., 2013b; Artetxe et al., 2018). Cross-lingual named-entity recognition for low-resource languages can also be ameliorated with the aid of dictionaries (Mayhew et al., 2017; Xie et al., 2018).

**Heterogeneous graph neural network** – GNNs encode each node in a graph to a vector by considering its attributes and the graph structure (Gori et al., 2005; Kipf and Welling, 2016; Veličković et al., 2017). They can be understood as message passing between nodes guided by edges to update the states of nodes (Gilmer et al., 2017; Hamilton et al., 2017). GNNs have been applied to many real-world problems, e.g., knowledge graph (Schlichtkrull et al., 2018), natural science (Sanchez-Gonzalez et al., 2018), and NLP (Yao et al., 2019). Heterogeneous GNN (HGNN) (Wang et al., 2019; Hu et al., 2020) is a type of GNNs that considers the types of nodes and edges in a heterogeneous graph that contains multiple types of nodes or edges.

## 3 Problem Formulation

In this section, we define our problem formulation. Let $\mathcal{X}$ be an input space and $\mathcal{Y}$ be an output space. Without loss of generality, we consider a document classification problem where $\mathcal{X}$ is a document space and $\mathcal{Y} = \{1, 2, \ldots, n_c\}$ is a set of classes, where $n_c$ denotes the number of classes. In this problem, we have languages $\mathcal{L} = \{\ell_0, \ell_1, \ldots, \ell_{n_s}\}$, where $\ell_0$ is the target language and $\ell_1, \ldots, \ell_{n_s}$ are source languages.

To define a word embedding function, let $\mathcal{V}^\ell$ be a vocabulary space and $d_\ell$ be the dimensionality of the word embeddings of a language $\ell$. Then, $E^\ell \colon \mathcal{V}^\ell \to \mathbb{R}^{d_\ell}$ is a word embedding function for a language $\ell$. Next, we define a bilingual dictionary $D^{\ell_i \to \ell_j} \colon \mathcal{V}^{\ell_i} \to 2^{\mathcal{V}^{\ell_j}}$ as a mapping from a known word in a language $\ell_i$ to a set of known words in a language $\ell_j$.

The following data are provided in this problem setting:

- Source word embeddings: $E^{\ell_1}, \ldots, E^{\ell_{n_s}}$.
- Bilingual dictionaries: $D^{\ell_1 \to \ell_0}, \ldots, D^{\ell_{n_s} \to \ell_0}$ and $D^{\ell_0 \to \ell_1}, \ldots, D^{\ell_0 \to \ell_{n_s}}$.
- Target labeled data: $X^{\mathrm{T}} := \{(\mathbf{x}_i, y_i)\}_{i=1}^n$ in a low-resource language $\ell_0$.

The goal of this problem is to learn a classifier $f \colon \mathcal{X} \to \mathcal{Y}$ that optimizes an evaluation metric of interest such as the accuracy or $F_1$-measure with respect to the target distribution.

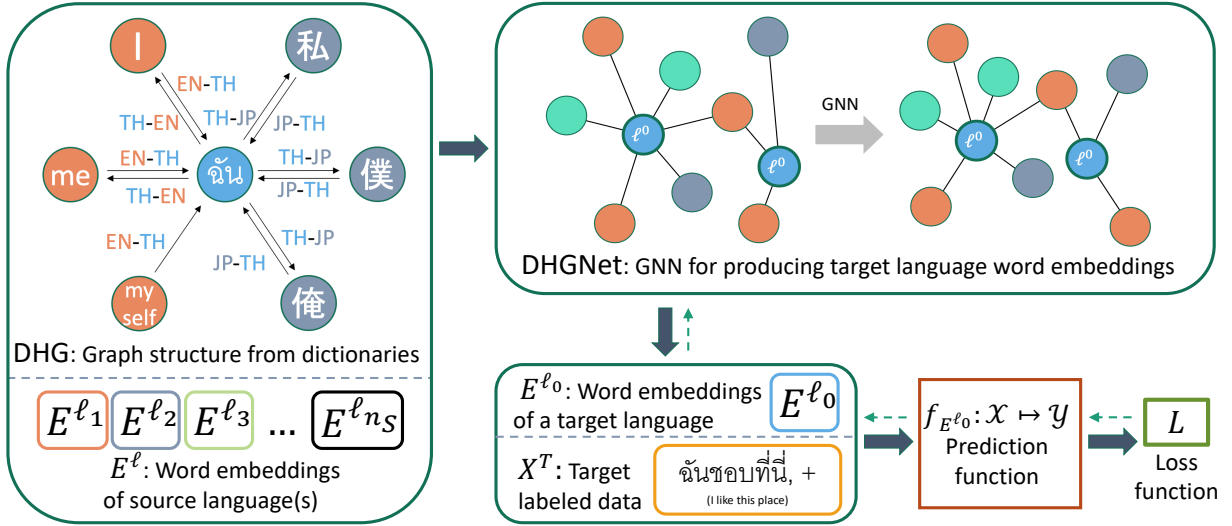

Figure 1: Overview of DHGNet. DHGNet takes DHG and word embeddings $E^{\ell}$ of source language(s) as inputs to produce word embeddings of the target language $E^{\ell_0}$. Then $E^{\ell_0}$ are used by a prediction function $f_{E^{\ell_0}}$ to predict labels and optimize the loss of a target task. The bold and dashed arrows indicate forward and backward passes, respectively. $E^{\ell_0}$, parameters of DHGNet, are trainable. $f_{E^{\ell_0}}$ can be a neural network. Colors indicate languages.

## 4 Proposed Method

In this section, we propose a novel method for cross-lingual transfer using source word embeddings and bilingual dictionaries. Figure 1 illustrates an overview of our proposed method.

### 4.1 Dictionary-based Heterogeneous Graph (DHG)

First, we use bilingual dictionaries and words in target labeled data to construct a heterogeneous graph. Let $\mathcal{V}^{\ell_0}$ be a vocabulary set of the target language, $\mathcal{V} = \bigcup_{i=0}^{n_S} \mathcal{V}^{\ell_i}$ be the vocabulary set of all languages of interest, and $\phi : \mathcal{V} \rightarrow \mathcal{L}$ be the word-to-language mapping[1]. DHG can be defined as follows.

**Definition 1** (Dictionary-based Heterogeneous Graph). *Dictionary-based heterogeneous graph $G$ is a directed graph $G = (\mathcal{V}, \mathcal{E}, \mathcal{L}, \phi)$ where $\mathcal{V}$ is the set of nodes (words), $\mathcal{E}$ is the set of edges: $\mathcal{E} = \bigcup_{\ell_i, \ell_j \in \mathcal{L}} \mathcal{E}^{\ell_i, \ell_j}$ where $\mathcal{E}^{\ell_i, \ell_j} = \{(v_1, v_2) | v_2 \in D^{\ell_i \rightarrow \ell_j}(v_1)\}$.*

An example of DHG can be observed from Figure 1. It is straightforward to see that DHG can be built by using words from target labeled data and their translations. Common words in the target language found in dictionaries can also be added.

---

[1]For notational simplicity, we assume words are non-overlap among languages. In the implementation, same words in different languages are treated as different nodes in DHG.

### 4.2 Dictionary-based Heterogeneous Graph Neural Network (DHGNet)

We propose DHGNets to combine the different semantic information of both word embeddings and DHG to obtain $d$-dimensional target word embeddings $E^{\ell_0} : \mathcal{V}^{\ell_0} \rightarrow \mathbb{R}^d$.

Intuitively, DHGNet works by exploiting the common structure between languages encoded in word embeddings. For example, a word "cat" in English can link to many co-occurrence words that helps in understanding the characteristics of real cats, and that can be shared to a low-resource for the classification task. The GNN in DHGNet (Figure 1) is vital to allow words of different languages to share information. That makes words from a low-resource inherits useful information from high-resources quantified by an attention mechanism, which will be introduced later.

Given the current target word embedding[2] $E^{\ell_0}$, DHG $G$, and source word embeddings $E^{\ell}$, we update the target word embedding with two steps: cross-lingual transformation and propagation with multi-source HGNN.

#### 4.2.1 Cross-lingual Transformation

Pretrained word embeddings of different source languages are usually trained separately. Thus, we should assume that they belong to different spaces and can have different dimensionalities, namely, $d_\ell$. For the primary step, we must transform them to a common space. To achieve that, we use the following mapping $E$ to map each word to the target word embedding space:

$$E(v) = \begin{cases} \mathbf{W}^{\phi(v)} E^{\phi(v)}(v) & \text{if } \phi(v) \neq \ell_0 \\ E^{\ell_0}(v) & \text{otherwise,} \end{cases}$$

where $\mathbf{W}^{\phi(v)} \in \mathbb{R}^{d \times d_\ell}$ is a trainable cross-lingual linear transformation.

---

[2]$E^{\ell_0}$ is randomly initialized at the first step of training.

### 4.2.2 Propagation with Multi-source HGNN

GNNs effectively learn node embedding via propagating and aggregating the node features (embeddings) based on the graph structure. Therefore, we can utilize a GNN to learn target word embeddings by effectively aggregating and synthesizing the embeddings from semantically related source words (neighboring nodes): $E \leftarrow \text{GNN}(E, G)$. One simple way is to use Graph Attention Networks (GATs) (Veličković et al., 2017) that can learn the important of each node in a graph. Particularly, for a *target node* $t \in \mathcal{V}$ in any language, GAT performs the following graph operation:

$$\mathbf{h}_s = \mathbf{W} E(s), \mathbf{h}_t = \mathbf{W} E(t)$$
$$\alpha_{s,t} = \underset{\forall s' \in N_G(t)}{\text{Softmax}} \left( \text{LeakyReLU}\left(\mathbf{a}^\top [\mathbf{h}_s \| \mathbf{h}_t]\right)\right),$$
$$\bar{\mathbf{h}}_t = \sigma\left(\sum_{\forall s \in N_G(t)} \alpha_{s,t} \mathbf{h}_s\right), \quad (1)$$

where $s \in \mathcal{V}$ and $\alpha \in \mathbb{R}$ denote *source node* and the attention score, respectively; $\mathbf{W} \in \mathbb{R}^{d_{out} \times d}$ and $\mathbf{a} \in \mathbb{R}^{2 d_{out}}$ are parameterized trainable weight matrix and vector of the layer, respectively; $N_G(t)$ denotes all in-neighbors of a node $t$; $\sigma$ denotes activation function; $\|$ is the concatenation operation. GATs usually use with *multi-head attention* by performing message passing on multiple $K$ independent heads, then concatenate the outputs: $\bar{\mathbf{h}}_t = \|_{k=1}^K \sigma\left(\sum_{\forall s \in N_G(t)} \alpha_{s,t}^k \mathbf{h}_s^k\right)$. Finally, $\bar{\mathbf{h}}_t$ is then used to update $E(t)$. In this paper, we use GELU as the activation function and set the output dimensions of each head $d_{out}$ to $d/K$. The number of GNN layers is set to 2.

Despite the ability of GAT to learn the saliency of nodes, the model lacks the awareness of language differences. This is because GAT is a *homogeneous* GNN that treats all nodes and edges identically regardless of their types, which may lead to suboptimal performance.

To effectively utilize DHG, we propose a multi-source HGNN consisting of two steps. The first step is to aggregate the relevant translations for each language pair (word-level aggregation), and then aggregate the knowledge over all language pairs (language-level aggregation). For example, a Thai word "ร้าน" can be translated to "shop" and "store" in English, and "店" (shop) and "餐厅" (restaurant) in Chinese. If a task is sentiment analysis of restaurant reviews, word-level aggregation decides to put more weight on "餐厅" than "店", while language-level aggregation decides to put more weight on Chinese-Thai than English-Thai language pair because "餐厅" is the most relevant meaning to the task.

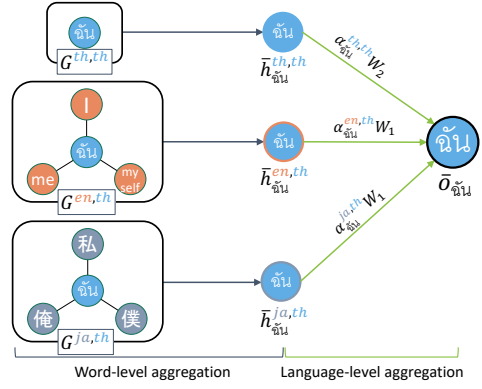

Figure 2: Example of aggregation by multi-source HGNN for the target node "ฉัน" in Thai language.

Mathematically, a homogeneous bilingual subgraph is defined as $G^{\ell_i, \ell_j} = \left(\mathcal{V}^{\ell_i} \cup \mathcal{V}^{\ell_j}, \mathcal{E}^{\ell_i, \ell_j}\right)$, where either $\ell_i$ or $\ell_j$ is the target language $\ell_0$.

**Word-level aggregation:** For any target node in any language $t \in \mathcal{V}$, *bilingual-specific node features* $\bar{\mathbf{h}}_t^{\ell_i, \ell_j}$ is calculated based on Eq. (1) using $G^{\ell_i, \ell_j}$ with trainable parameters $\mathbf{W}^{\ell_i, \ell_j}$ and $\mathbf{a}^{\ell_i, \ell_j}$.

**Language-level aggregation:** Given $\bar{\mathbf{h}}_t^{\ell_i, \ell_j}$ from word-level aggregation and let $\ell = \phi(t)$ be the language of $t$. With trainable parameters $\mathbf{W}_1, \mathbf{W}_2$ and $\mathbf{a}_1$, the target output for $t$ can be calculated by

$$\mathbf{o}_t^r = \mathbf{W}_1 \bar{\mathbf{h}}_t^r, \mathbf{o}_t^{\ell, \ell} = \mathbf{W}_2 E(t)$$
$$\alpha_t^r = \underset{\forall r': |N_{G^{r'}}(t)| > 0}{\text{Softmax}}\left(\text{LeakyReLU}\left(\mathbf{a}_1^\top [\mathbf{o}_t^r \| \mathbf{o}_t^{\ell, \ell}]\right)\right)$$
$$\bar{\mathbf{o}}_t = \sigma\left(\sum_{\forall r: |N_{G^r}(t)| > 0} \alpha_t^r \mathbf{o}_t^r\right),$$

where $r$ denotes a pair of languages including a pair of the same language. Figure 2 illustrates an example of aggregation by multi-source HGNN. Note that $\mathbf{W}_1$ is the same across language pairs and $\mathbf{W}_2$ is the same for all languages. Finally, multi-head attention mechanism can also be applied to the tar-get output $\bar{\mathbf{o}}_t$. To avoid oversmoothing when training GNN (Li et al., 2018), we also add a residual connection (Chen et al., 2020) before updating the embedding: $E(t) \leftarrow \bar{\mathbf{o}}_t + E(t)$ for each layer.

### 4.2.3 End-to-end Optimization

To simultaneously train our DHGNet and a prediction function, we use the target word embedding $E^{\ell_0}$ as an input of the prediction function and back-propagate a loss to update the trainable parameters of DHGNet, as illustrated in Figure 1.

Let us define $f_{E^{\ell_0}}$ a prediction function that uses target word embeddings to extract features of the input space $\mathcal{X}$. Also, we define $\Theta_{\text{DHG}}$ to be trainable parameters for DHGNet and $L$ to be a loss function to train a classifier $f_{E^{\ell_0}}$. For example, $L$ can be the average cross-entropy loss over training data. Then, the gradient information for the DHGNet can be obtained by the following simple chain rule: $\frac{\partial L}{\partial \Theta_{\text{DHG}}} = \frac{\partial L}{\partial E^{\ell_0}} \frac{\partial E^{\ell_0}}{\partial \Theta_{\text{DHG}}}$. Therefore, $L$ is used to simultaneously train the prediction function $f_{E^{\ell_0}}$ and DHGNet to produce good task-specific word embeddings $E^{\ell_0}$ for solving the target task.

It is worth pointing out that in the cross-lingual transformation step, one can learn $\mathbf{W}^{\phi(v)}$ to control the behavior of an embedding function $E$. For example, using contrastive learning (Lazaridou et al., 2015; Joulin et al., 2018) can encourage $E$ to produce similar outputs between a pair of nodes that contain translation between each other.

## 5 Experiments

In this section, we conducted experiments to compare our method with baselines and analyze the method under different conditions.

### 5.1 Experiment Setup

| Setting | $\ell_0$ | Task | # docs | # classes |
|---|---|---|---|---|
| Bengali | bn | Topic | 14,126 | 6 |
| Bosnian | bs | Sentiment | 7,241 | 2 |
| Malayalam | ml | Topic | 6,000 | 4 |
| Tamil | ta | Topic | 11,700 | 3 |
| Thai-T | th | Intent | 16,175 | 7 |
| Thai-W | th | Rating | 39,995 | 5 |

Table 1: Details and statistics of each setting.

**Dataset** – We conducted experiments on four datasets in five different languages. For Thai language (th), we used Truevoice (Thai-T) and Wongnai (Thai-W) datasets in PyThaiNLP library (Phatthiyaphaibun et al., 2016). We used news articles in Bengali (bn), Malayalam (ml), and Tamil (ta) languages from IndicNLPSuite (Kakwani et al., 2020). For European languages, we used Twitter sentiment (Mozetič et al., 2016) in Bosnian (bs) language. In total, we have six different settings. We used the average accuracy and macro-average $F_1$-measure of five runs as evaluation metrics. We provide the statistics of each setting in Table 1.

**Dictionary** – We used word2word (Choe et al., 2020). The bilingual dictionaries were automatically constructed from OpenSubtitles2018 dataset (Lison et al., 2018).

**Pretrained word embedding of DHGNet** – We used publicly available word vectors from Fast-Text[3] (Bojanowski et al., 2017). For source languages, we used English (en), Arabic (ar), Chinese (zh), French (fr), Persian (fa), and Spanish (es).

### 5.2 Comparison with baselines

We conducted experiments using multiple baseline methods that can potentially apply to our settings, including a statistical method, task-independent word embeddings, and pretrained models. The list is as follows:

- SVM: a Support Vector Machine classifier that takes Term Frequency-Inverse Document Frequency features as an input.
- FastText (Bojanowski et al., 2017): a task-independent word embedding trained on Wikipedia and Common Crawl only of a target language.
- RCSLS (Joulin et al., 2018): a CLWE trained on Wikipedia. We used public FastText[4] version.
- USE (Yang et al., 2020): a multilingual model pretrained by solving multiple tasks, including question answer, natural language inference, and translation ranking in 16 languages. USE is available only for Thai language in our experiments. Data sources include online forums, QA websites, and public datasets.
- ULMFiT (Howard and Ruder, 2018): a pretrained language model pretrained on Wikipedia articles. We used a publicly available pretrained model of a target language.
- mBERT (Devlin et al., 2019): a multilingual pretrained BERT model trained on 104 languages. The model is trained on machine-translated versions of BookCorpus (Zhu et al., 2015) and Wikipedia articles.

---

[3]https://fasttext.cc/docs/en/pretrained-vectors.html
[4]https://fasttext.cc/docs/en/aligned-vectors.html

|  | Bengali | | Bosnian | | Malayalam | | Tamil | | Thai-T | | Thai-W | |
|---|---|---|---|---|---|---|---|---|---|---|---|---|
|  | Acc | $F_1$ | Acc | $F_1$ | Acc | $F_1$ | Acc | $F_1$ | Acc | $F_1$ | Acc | $F_1$ |
| SVM | 84.05 | 72.29 | 62.73 | 66.42 | 89.17 | 89.40 | 96.75 | 96.78 | 30.93 | 8.27 | 46.41 | 18.94 |
| FastText | 83.27 | 80.57 | 68.94 | 71.34 | 90.67 | 90.85 | 96.67 | 96.69 | 75.00 | 78.62 | 48.17 | 16.13 |
| FastText-LSTM | 94.26 | 91.05 | 74.40 | 75.67 | 92.83 | 93.09 | 98.03 | 98.05 | 84.43 | 86.83 | 59.27 | 45.16 |
| RCSLS | 83.70 | 78.57 | 70.05 | 72.14 | - | - | 96.32 | 96.36 | 30.81 | 8.76 | 48.17 | 16.13 |
| RCSLS-LSTM | 93.76 | 90.30 | 70.53 | 73.30 | - | - | 98.03 | 98.04 | 84.61 | 86.82 | 59.78 | 46.72 |
| USE | - | - | - | - | - | - | - | - | 84.77 | 87.45 | 50.83 | 21.01 |
| ULMFiT | 93.62 | 91.10 | - | - | 91.50 | 91.74 | 98.03 | 98.04 | 84.61 | 85.49 | 59.18 | 46.71 |
| mBERT | 83.40 | 76.52 | 75.76 | 77.21 | 91.60 | 91.80 | 97.01 | 97.02 | 87.36 | 88.67 | 46.20 | 21.76 |
| mBERT-LSTM | 89.23 | 84.91 | 77.16 | 78.52 | 91.17 | 91.44 | 96.50 | 96.52 | 87.05 | 88.08 | 46.37 | 12.67 |
| XLM-R | 94.54 | 92.44 | 53.07 | 69.34 | **95.00** | **95.14** | 97.44 | 97.45 | **89.96** | **91.63** | 46.37 | 12.67 |
| no-DHGNet | 93.05 | 89.16 | 69.15 | 70.83 | 91.17 | 91.44 | 97.78 | 97.80 | 85.64 | 87.34 | 59.24 | 46.04 |
| **DHGNet**$_{en}$ | 94.26 | 91.08 | 74.52 | 76.35 | 92.43 | 92.67 | 98.12 | 98.13 | 89.54 | 91.28 | 60.42 | 49.13 |
| **DHGNet**$_{multi}$ | **95.18** | **92.67** | **78.05** | 78.37 | 93.50 | 93.67 | **98.29** | **98.30** | 88.74 | 90.34 | 60.10 | **51.29** |

Table 2: Text classification results (mean accuracy and macro-average $F_1$-measure). The bolded numbers indicate the best performance in each case. The underlined numbers indicate the second best performances. The hyphen indicates the unavailability of a pretrained model in a target language.

- XLM-R (Conneau et al., 2020): a large multilingual model pretrained on colossal-scale corpus with multi-objective pretraining (Conneau and Lample, 2019).
- DHGNet: We used an AWD-LSTM (Merity et al., 2018) as the prediction function. **DHGNet**$_{en}$ and **DHGNet**$_{multi}$ denote DHGNet with English and multiple source languages (ar,en,es,fa,fr,zh), respectively. We also included **no-DHGNet** that is an AWD-LSTM without DHGNet for a comparison.

For FastText, RCSLS, and USE, we used SVM, random forest, and logistic regression as classification methods, and reported the result only of the one with the best validation performance to save space. For a better comparison, we also used the AWD-LSTM as the classifier for FastText, RCSLS, and mBERT named FastText-LSTM, RCSLS-LSTM, and mBERT-LSTM. All pretrained models are fine-tuned on a target data. For implementation details and hyperparameter settings, please see Appendix A.

The results are listed in Table 2. We observe that DHGNet outperformed no-DHGNet in most settings from 1.18% in Thai-W to significantly 8.90% in Bosnian in absolute accuracy.

Furthermore, apart from XLM-R, DHGNet also achieved higher performance than other pretrained baselines, even though it did not access to any large corpus of those target languages, such as Wikipedia data. Note that XLM-R requires much higher computational cost[5] and has different input source (Wikipedia data in 100 languages and more) compared with DHGNet (source word embeddings, bilingual dictionaries).

Figure 3 shows the results with varying training size. It can be observed that DHGNet can still perform relatively well under high data scarcity. The gap between DHGNet and no-DHGNet is larger as the training size decreases. Moreover, in Bosnian, DHGNet$_{multi}$ consistently outperformed DHGNet$_{en}$ and other methods.

The performance of all methods also depends on the data provided for them. To the best of our knowledge, there does not exist any methods that use the same input source as our proposed DHGNets. As a result, it is difficult to provide a fairness comparison for all methods due to different data accessibility. For example, Wikipedia data of the target language were used by all baselines for pretraining or producing word embeddings. On the other hand, DHGNets never observe such data but use bilingual dictionaries to construct DHG. Nevertheless, our experiment shows that using bilingual dictionaries and source word embeddings can construct a classifier that is competitive to baselines accessing to extremely large corpora. Thus, we believe that DHGNet is the current best solution if parallel corpora cannot be obtained but only bi-lingual dictionary.

### 5.3 Analysis

We conducted further experiments to analyze the effect of (1) quality of dictionary, (2) source language, (3) source word embeddings, and (4) choice of graph neural networks to the classification performance of DHGNet.

---
[5]XML-R required 40GB of GPU memory in our experiments to finetune compared to maximally 16GB used by others.

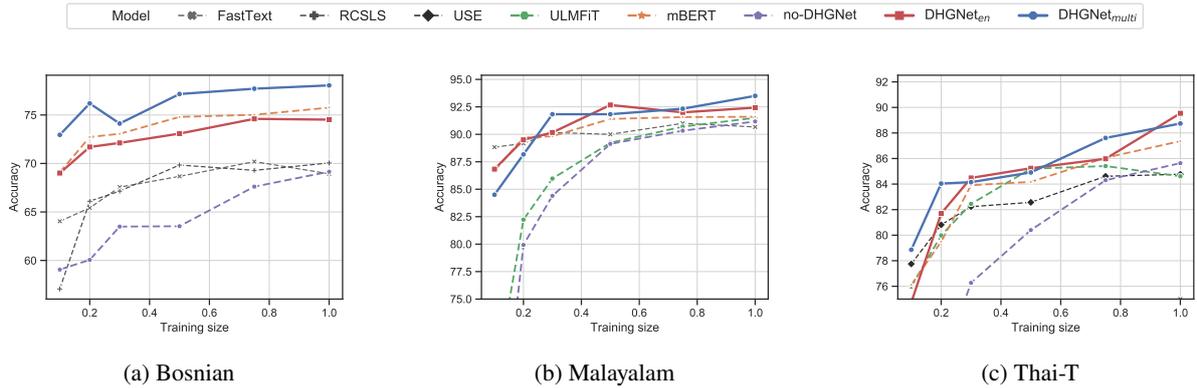

(a) Bosnian  (b) Malayalam  (c) Thai-T

Figure 3: Classification results (mean accuracy) on each setting with varying training size.

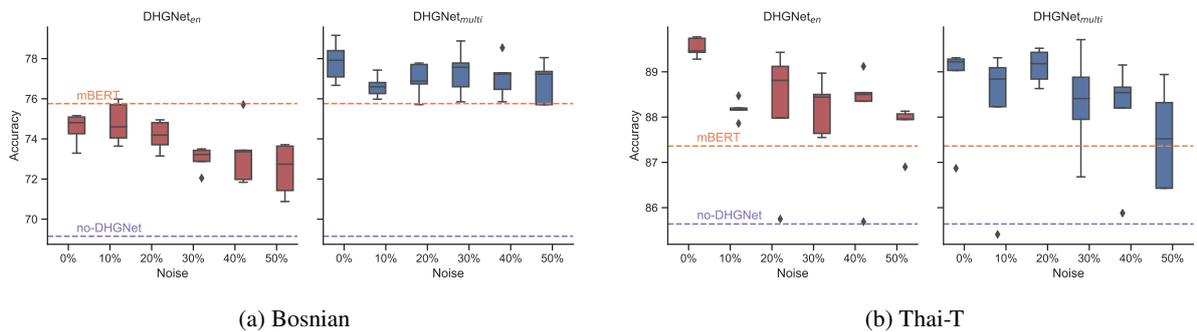

(a) Bosnian  (b) Thai-T

Figure 4: Classification results on each data with varying degree of noise (added incorrect translation) in dictionary. Both DHGNets could maintain their performances and beat baselines even the dictionary were heavily tampered.

### 5.3.1 Quality of Dictionary

In practice, auto-generated dictionaries have certain noises for high co-occurrence words and wrong word-segmentation as we found from word2word. However, such noises are difficult to synthesize. To analyze the effect of the quality of dictionaries, we injected noise by randomly adding incorrect translation to worsen the quality of dictionary. For example, "猫" (cat) might have been added to the translation of "dog". Random noise should worsen the quality of dictionaries than practical noises because the added words are likely to be highly dissimilar to the correct translation.

Figure 4 shows the effect of dictionary qualities. Both DHGNets could maintain performance even in high noise presence and surpassed mBERT. Given 50% noise rate, DHGNets did not suffer from negative transfer because they achieved superior performance than that of no-DHGNet.

DHGNets achieved high level of dictionary robustness owing to being able to identify the suitable translations of each target word. Translations that benefit the task were given more attention, while reducing neutral and harmful ones. Knyazev et al.

| $\ell$ | Bengali | Bosnian | Malayalam | Thai-T |
|---|---|---|---|---|
| ar | 93.55 (6) | 74.85 (4) | 92.90 (2) | 87.68 (4) |
| en | 94.26 (3) | 74.52 (5) | 92.43 (5) | **89.54 (1)** |
| es | 93.98 (5) | **75.60 (1)** | 92.67 (4) | 88.86 (3) |
| fa | **94.68 (1)** | 74.20 (6) | 92.17 (6) | 89.02 (2) |
| fr | 94.05 (4) | 75.03 (3) | **92.93 (1)** | 86.85 (5) |
| zh | 94.54 (2) | 75.32 (2) | 92.83 (3) | 86.34 (6) |

Table 3: Accuracy of DHGNet with different source languages. The numbers in parentheses indicate ranking.

(2019) showed that the attention mechanism in GNNs can improve robustness to noisy graphs by attending to important and avoiding noisy parts. By having more sources, $DHGNet_{multi}$ outperformed $DHGNet_{en}$ in most cases.

### 5.3.2 Choice of Source Languages

We investigate the effect of choosing different source languages. The results are listed in Table 3. We observe that the choice of source languages can improve up to 3% absolute accuracy. There is no best source language for all settings. English, which is the most resource-rich, could give inferior performance in some cases. The worst performance of each setting in Table 3 still outperformed no-DHGNet in Table 2, indicating no negative transfer.

### 5.3.3 Quality of Source Word Embeddings

To analyze the quality of source word embeddings, we changed the source word embeddings of DHGNet$_{en}$ to GloVe (Pennington et al., 2014), RCSLS, and Wiki2vec (Yamada et al., 2016). We also included a variant that does not use any pretrained word embeddings by replacing them with trainable randomly-initialized word vectors (RandInit).

Table 4 shows the performance of DHGNet with different source word embeddings. RandInit can be observed to perform worst. This indicates the benefit of using source word embeddings.

|  | Bengali | Bosnian | Malayalam | Thai-T |
|---|---|---|---|---|
| RandInit | 94.40 (3) | 71.57 (5) | 91.33 (5) | 87.24 (4) |
| FastText | 94.33 (4) | **76.60 (1)** | 92.43 (4) | **89.54 (1)** |
| GloVe | 94.05 (5) | 75.91 (2) | **93.50 (1)** | 87.89 (3) |
| RCSLS | 94.68 (2) | 75.50 (3) | 92.83 (3) | 88.14 (2) |
| Wiki2vec | **95.04 (1)** | 75.29 (4) | **93.50 (1)** | 87.17 (5) |

Table 4: Accuracy of DHGNet$_{en}$ with different source pretrained word embeddings. The numbers in parentheses indicate ranking.

### 5.3.4 Choice of Graph Neural Networks

One core component of DHGNet is Multi-source HGNN described in Section 4.2.2. To analyze its effect in the multi-source scenario, we replaced our proposed multi-source HGNN in DHGNet$_{multi}$ to other GNNs. We included homogeneous GNNs: GCN (Kipf and Welling, 2016) and GAT (Veličković et al., 2017) and heteregeneous GNN: RGCN (Schlichtkrull et al., 2018). RGCN employs distinct weights for each relation (language pair) with the mean aggregation. Note that to use a different GNN to solve the problem, we must still use our proposed DHG (Section 4.1), cross-lingual transformation technique (Section 4.2.1), and the end-to-end optimization (Section 4.2.3). Table 5 lists the performance of DHGNet$_{multi}$ with different GNNs. It can be observed that our proposed method outperformed other baseline GNNs.

|  | Bengali | Bosnian | Malayalam | Thai-T |
|---|---|---|---|---|
| GCN | 94.61 (2) | 77.28 (4) | 92.83 (2) | **89.06 (1)** |
| GAT | 94.54 (3) | 77.43 (2) | 91.33 (4) | 87.86 (4) |
| RGCN | 94.12 (4) | 77.32 (3) | 92.17 (3) | **89.06 (1)** |
| Ours | **95.18 (1)** | **78.05 (1)** | **93.50 (1)** | 88.74 (3) |

Table 5: Accuracy of DHGNet$_{multi}$ with different GNNs. The numbers in parentheses indicate ranking.

## 6 Discussion and Conclusions

We proposed a method based on heterogeneous graph neural networks called DHGNet to transfer the source word embeddings through bilingual dictionaries. Without task-specific source data, DHGNet demonstrates that it can outperform models pretrained on extremely large corpora. Furthermore, our results revealed that DHGNet can perform well even though dictionaries contain many incorrect translations. Its robustness opens the possibility to use a wider range of dictionaries such as an automatically constructed dictionary and crowdsourced dictionary.

**Limitation** – Because our method operates on the word level, some words may not be found in bilingual dictionaries although they are in the dictionaries in a different form. This may occur in languages that have declensions, e.g., Malayalam, Japanese, and Sanskrit. For computational limitation, DHGNet$_{multi}$ can be trained with a DHG upto 30K target words[6], 110K source words, and 1M edges for long documents, e.g., news articles, on a single NVIDIA Tesla V100-16GB.

### Acknowledgments

This work was supported by JST CREST (Grant Number JPMJCR1687), JSPS Grant-in-Aid for Scientific Research (Grant Number 17H01785, 21K12042), and the New Energy and Industrial Technology Development Organization (Grant Number JPNP20006). Nuttapong Chairatanakul was supported by MEXT scholarship. Nontawat Charoenphakdee was supported by MEXT scholarship and Google PhD Fellowship Program.

---

[6]Only words found in bilingual dictionaries are counted.

# A Implementation Notes and Hyperparameter Settings

We implemented DHGNet based on PyTorch[7] and PyTorch Geometric (Fey and Lenssen, 2019) libraries in Python 3.6. To train AWD-LSTM, we used fastai (Howard and Gugger, 2020) library version 1.0.61. All experiments except for XLM-R were conducted on Intel Xeon Gold 6148 CPU @ 2.4 GHz, 5 Cores, 60GB RAM, NVIDIA Tesla V100-16GB. The experiments for XLM-R were conducted on Intel Xeon Platinum 8360Y CPU @ 2.4 GHz, 9 Cores, 60GB RAM, NVIDIA A100-40GB.

We obtained RCSLS word embeddings from publicly available FastText[8] version. For FastText and RCSLS, we averaged the word embeddings for each input data to form the input features, then use them as input to train and predict with a classifier. The classifier is selected from {Logistic, RBF-SVM, Random Forest} performing best on the validation set using Scikit-learn library (Pedregosa et al., 2011) with their default hyperparameter settings. The similar was also applied to USE.

For ULMFiT, we obtained the pretrained model in each target language from: iNLTK (Arora, 2020) for Bengali, Malayalam, and Tamil languages and PyThaiNLP library (Phatthiyaphaibun et al., 2016) for Thai language.

For mBERT and XLM-R, we finetuned it on labeled target data using simpletransformers[9] library with AdamW optimizer, learning rate: $4e-5$, batch size: 32, maximum sequence length: 300, and the number of epochs: 15.

For models with AWD-LSTM, including both DHGNets, we performed language model training on the input documents of a target language in the training set before classification learning. We experimentally found that creating language models with the next word prediction task before training for the target task yields better results than without it in most cases. Saunshi et al. (2021) justify that the next word prediction tasks pretrain the models in a meaningful way for the target classification tasks. The models shared the same AWD-LSTM hyperparameters for each experimental setting. The hyperpameters were derived from the default setting in fastai library:

[7] https://pytorch.org/
[8] https://fasttext.cc/docs/en/aligned-vectors.html
[9] https://github.com/ThilinaRajapakse/simpletransformers

- embedding dropout: 0.02 and 0.05
- the number of LSTM cells: 1152 and 1152
- LSTM input dropout: 0.25 and 0.4
- LSTM weight dropout: 0.2 and 0.5
- between LSTMs dropout: 0.15 and 0.3
- output dropout: 0.1 and 0.4

for language model training and classification training, respectively. We set the embedding dimension $d$ to 300. The number of LSTM layers was set to 1 for Albanian and Bosnian settings because the texts are short, while it was set to 3 for the others. Batch size, maximum sequence length, and BPTT length were set to 32, 700, and 70, respectively. For Thai language, we used the default tokenizer and text pre/post-processing from PyThaiNLP. Otherwise, we used the default from fastai.

For DHGNet, we set the number of GNN layers to 2. The number of multi-heads and output dimensions of each head $d_{out}$ were set to 10 and 30, respectively. We applied Layer Normalization (Ba et al., 2016) to each GNN layer's input and the output of DHGNet. For Albanian and Bosnian settings, we added 30,000 common words found in bilingual dictionaries for handling the Out-of-Vocabulary problem in the test set. We also used contrastive learning as mentioned in the last paragraph of Section 4.2.3 for refining the initial $E^{\ell_0}$ and the cross-lingual transformation in these two settings.